\documentclass{article}

\usepackage{graphicx} 
\usepackage{subfigure}

\usepackage{natbib,amsmath}

\usepackage{algorithm}
\usepackage{algorithmic}

\usepackage{hyperref}

\usepackage[accepted]{icml2013}

\def\BE{\vspace{-0.0mm}\begin{equation}}
\def\EE{\vspace{-0.0mm}\end{equation}}
\def\BEA{\vspace{-0.0mm}\begin{eqnarray}}
\def\EEA{\vspace{-0.0mm}\end{eqnarray}}

\newcommand{\fig}[1]{Fig.~\ref{fig:#1}}
\newcommand{\tab}[1]{Table~\ref{tab:#1}}
\newcommand{\secc}[1]{Section~\ref{sec:#1}}
\def\etal{{\textit{et~al.~}}}

\icmltitlerunning{Visualizing and Understanding Convolutional Networks}

\begin{document}

\twocolumn[
\icmltitle{Visualizing and Understanding Convolutional Networks}

\icmlauthor{Matthew D. Zeiler}{zeiler@cs.nyu.edu}
\icmladdress{Dept. of Computer Science, Courant Institute,
            New York University}
\icmlauthor{Rob Fergus}{fergus@cs.nyu.edu}
\icmladdress{Dept. of Computer Science, Courant Institute,
            New York University}

\icmlkeywords{boring formatting information, machine learning, ICML}

\vskip 0.3in
]

\begin{abstract}
Large Convolutional Network models have recently demonstrated
impressive classification performance on the ImageNet benchmark
\cite{Kriz12}. However there is no clear understanding of
why they perform so well, or how they might be improved. In this
paper we address both issues. We introduce a novel visualization
technique that gives insight into the function of intermediate feature
layers and the operation of the classifier. Used in a diagnostic role, these
visualizations allow us to find model architectures that outperform Krizhevsky
\etal on the ImageNet classification benchmark. We also perform an
ablation study to discover the performance contribution from different
model layers. We show our ImageNet model
generalizes well to other datasets: when the softmax classifier is
retrained, it convincingly beats the current state-of-the-art
results on Caltech-101 and Caltech-256 datasets.
\end{abstract}

\vspace{-4mm}
\section{Introduction}
\vspace{-2mm}
Since their introduction by \cite{Lecun1989} in the early 1990's, Convolutional
Networks (convnets) have demonstrated excellent performance at
tasks such as hand-written digit classification and face
detection. In the last year, several papers have shown that they can also deliver
outstanding performance on more challenging visual classification tasks. \cite{Ciresan12} demonstrate state-of-the-art performance on NORB and
CIFAR-10 datasets. Most notably, \cite{Kriz12} show
record beating performance on the ImageNet 2012 classification
benchmark, with their convnet model achieving an error rate of 16.4\%,
compared to the 2nd place result of 26.1\%. Several factors are
responsible for this renewed interest in convnet models: (i) the
availability of much larger training sets, with millions of labeled
examples; (ii) powerful GPU implementations, making the training of
very large models practical and (iii) better model regularization
strategies, such as Dropout \cite{Hinton12}.

Despite this encouraging progress, there is still little insight into
the internal operation and behavior of these complex models, or how
they achieve such good performance. From a scientific standpoint,
this is deeply unsatisfactory. Without clear understanding of how and
why they work, the development of better models is reduced to
trial-and-error. In this paper we introduce a visualization technique
that reveals the input stimuli that excite individual feature maps at any
layer in the model. It also allows us to observe the evolution of
features during training and to diagnose potential problems with the
model. The visualization technique we propose uses a multi-layered
Deconvolutional Network (deconvnet), as proposed by \cite{Zeiler11},
to project the feature activations back to the input pixel space. We also perform a
sensitivity analysis of the classifier output by occluding portions of
the input image, revealing which parts of the scene are important for
classification.

Using these tools, we start with the architecture of \cite{Kriz12} and
explore different architectures, discovering ones that outperform
their results on ImageNet. We then explore the generalization ability
of the model to other datasets, just retraining the softmax classifier
on top. As such, this is a form of supervised pre-training, which
contrasts with the unsupervised pre-training methods popularized by
\cite{Hinton2006a} and others \cite{Bengio2007,Vincent2008}. The
generalization ability of convnet features is also explored in
concurrent work by \cite{Donahue13}.

\vspace{-0mm}
\subsection{Related Work}
\vspace{-0mm} Visualizing features to gain intuition about the network
is common practice, but mostly limited to the 1st layer where
projections to pixel space are possible. In higher layers this is not
the case, and there are limited methods for interpreting
activity. \cite{Erhan09} find the optimal stimulus for each unit by
performing gradient descent in image space to maximize the unit's
activation. This requires a careful initialization and does not give
any information about the unit's invariances. Motivated by the
latter's short-coming, \cite{Le10} (extending an idea by
\cite{Berkes06}) show how the Hessian of a given unit may be computed
numerically around the optimal response, giving some insight into
invariances. The problem is that for higher layers, the invariances
are extremely complex so are poorly captured by a simple quadratic
approximation. Our approach, by contrast, provides a non-parametric
view of invariance, showing which patterns from the training set
activate the feature map. \cite{Donahue13} show
visualizations that identify patches within a dataset that are
responsible for strong activations at higher layers in the model. Our
visualizations differ in that they are not just crops of input images,
but rather top-down projections that reveal structures within each
patch that stimulate a particular feature map.

\section{Approach}

We use standard fully supervised convnet models throughout
the paper, as defined by \cite{Lecun1989} and \cite{Kriz12}. These models map a color 2D input image $x_i$,
via a series of layers, to a
probability vector $\hat{y_i}$ over the $C$ different classes.  Each
layer consists of (i) convolution of the previous layer output (or, in
the case of the 1st layer, the input image) with a set of learned
filters; (ii) passing the responses through a rectified linear
function ({\em $relu(x)=\max(x,0)$}); (iii)
[optionally] max pooling over local neighborhoods and (iv) [optionally]
a local contrast operation that normalizes the responses across
feature maps. For more details of these operations, see \cite{Kriz12} and \cite{Jarrett2009}. The top few
layers of the network are conventional fully-connected networks and
the final layer is a softmax classifier. \fig{arch} shows
the model used in many of our experiments.

We train these models using a large set of $N$ labeled images
$\{x,y\}$, where label $y_i$ is a discrete variable indicating the
true class. A cross-entropy loss function, suitable for image
classification, is used to compare $\hat{y_i}$ and $y_i$. The
parameters of the network (filters in the convolutional layers, weight
matrices in the fully-connected layers and biases) are trained by
back-propagating the derivative of the loss with respect to the
parameters throughout the network, and updating the parameters via
stochastic gradient descent. Full details of training are given in
\secc{training}.

\subsection{Visualization with a Deconvnet}
Understanding the operation of a convnet requires interpreting the
feature activity in intermediate layers. We present a novel way to
{\em map these activities back to the input pixel space}, showing what
input pattern originally caused a given activation in the feature
maps.  We perform this mapping with a Deconvolutional Network
(deconvnet) \cite{Zeiler11}. A deconvnet can be thought of as a
convnet model that uses the same components (filtering, pooling) but
in reverse, so instead of mapping pixels to features does the
opposite. In \cite{Zeiler11}, deconvnets were proposed as a way of
performing unsupervised learning. Here, they are not used in any
learning capacity, just as a probe of an already trained convnet.

\begin{figure}[h!]
\vspace{-3mm}
\begin{center}
\includegraphics[width=3.3in]{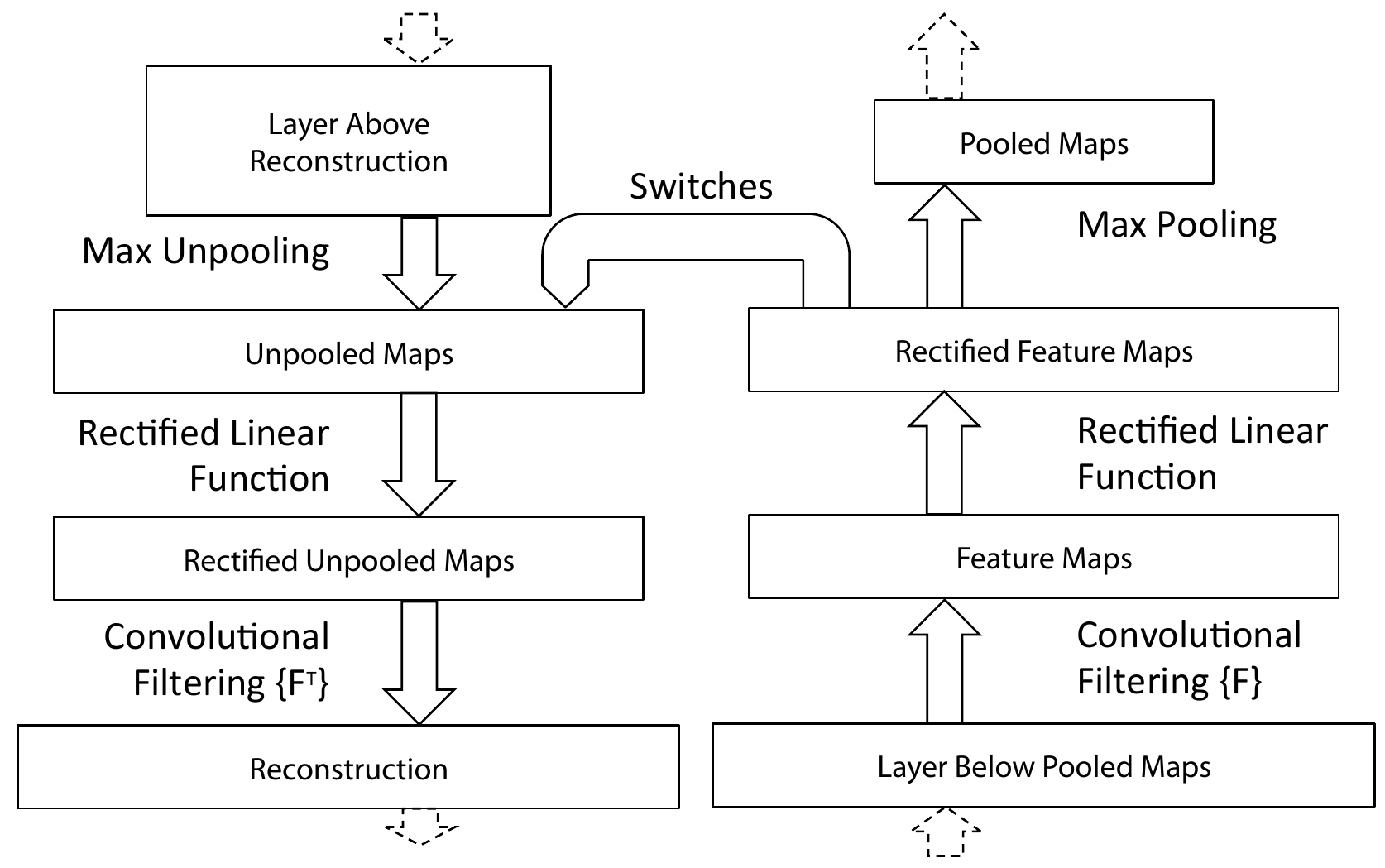}
\includegraphics[width=3.3in]{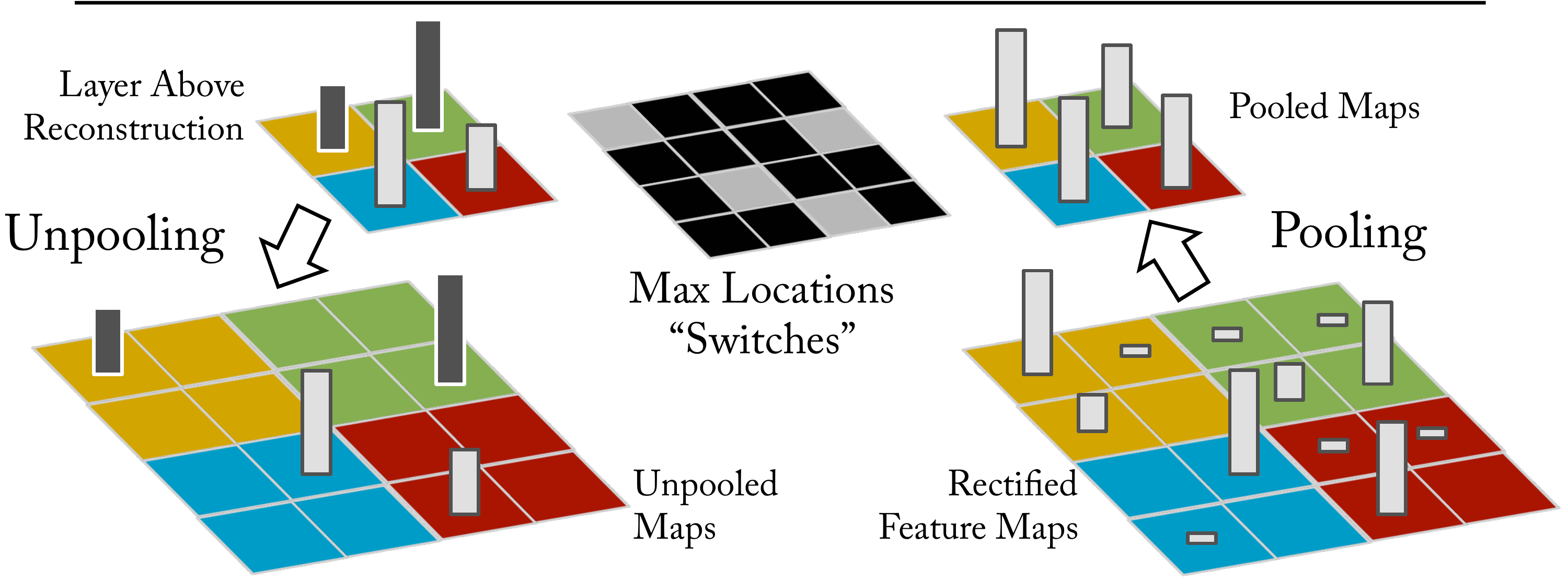}
\end{center}
\vspace*{-0.3cm}
\caption{Top: A deconvnet layer (left) attached to a convnet layer
  (right). The deconvnet will reconstruct an approximate version of
  the convnet features from the layer beneath. Bottom:
  An illustration of the unpooling operation in the deconvnet, using {\em switches}
  which record the location of the local max in each pooling region
  (colored zones) during pooling in the convnet. }
\label{fig:deconv}
\vspace*{-0.3cm}
\end{figure}

To examine a convnet, a deconvnet is attached to each of its layers,
as illustrated in \fig{deconv}(top), providing a continuous
path back to image pixels. To start, an input image is
presented to the convnet and features computed throughout the
layers. To examine a given convnet activation, we set all other activations in
the layer to zero and pass the feature maps as input to the attached
deconvnet layer. Then we successively (i) unpool, (ii) rectify and
(iii) filter to reconstruct the activity in the layer beneath that
gave rise to the chosen activation. This is then repeated until input
pixel space is reached.

\noindent {\bf Unpooling:} In the convnet, the max pooling operation
is non-invertible, however we can obtain an approximate inverse by
recording the locations of the maxima within each pooling region in a
set of {\em switch} variables. In the deconvnet, the unpooling
operation uses these switches to place the reconstructions from the
layer above into appropriate locations, preserving the structure of
the stimulus. See \fig{deconv}(bottom) for an illustration of the procedure.

\noindent {\bf Rectification:} The convnet uses {\em relu}
non-linearities, which rectify the feature maps thus ensuring the
feature maps are always positive. To obtain valid feature
reconstructions at each layer (which also should be positive), we pass
the reconstructed signal through a {\em relu} non-linearity.

\noindent {\bf Filtering:} The convnet uses learned filters to
convolve the feature maps from the previous layer. To invert this, the
deconvnet uses transposed versions of the same filters, but applied to the rectified maps,
not the output of the layer beneath. In
practice this means flipping each filter vertically and horizontally.

Projecting down from higher layers uses the switch settings generated
by the max pooling in the convnet on the way up. As these switch
settings are peculiar to a given input image, the reconstruction
obtained from a single activation thus resembles a small piece of the
original input image, with structures weighted according to their
contribution toward to the feature activation. Since the model is
trained discriminatively, they implicitly show which parts of the
input image are discriminative. Note that these projections are {\em
  not} samples from the model, since there is no generative process involved.

\section{Training Details} \label{sec:training}
\begin{figure*}[t!]
\vspace*{-0.2cm}
\begin{center}
\includegraphics[width=7in]{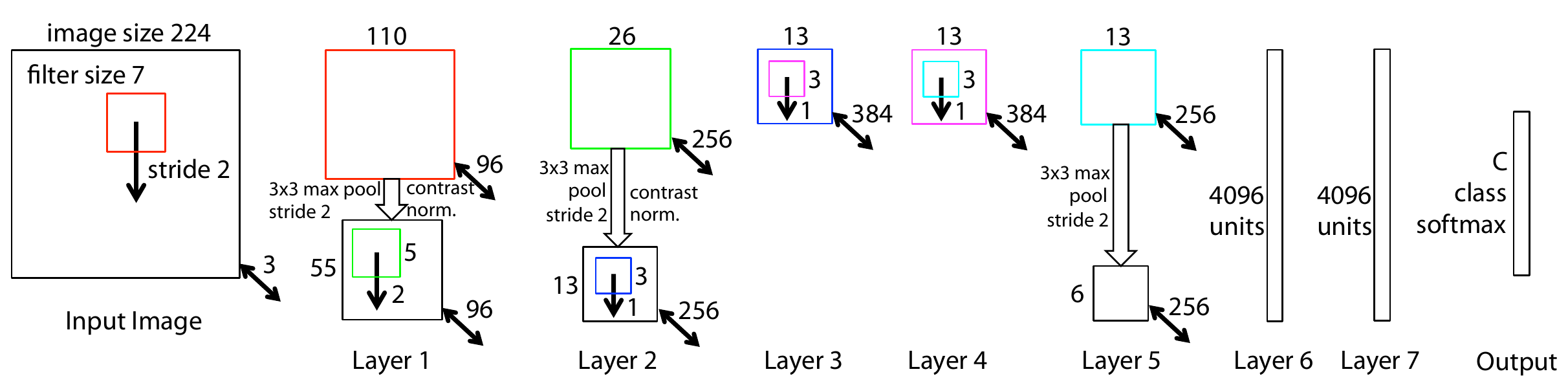}
\end{center}
\vspace*{-0.4cm}
\caption{Architecture of our 8 layer convnet model. A 224 by 224 crop of an
  image (with 3 color planes) is presented as the input. This is
  convolved with 96 different 1st layer filters (red), each of size 7 by 7,
  using a stride of 2 in both x and y. The resulting feature maps are
  then: (i) passed through a rectified linear function (not shown),
  (ii) pooled (max
  within 3x3 regions, using stride 2) and (iii) contrast normalized
  across feature maps to give 96 different 55 by 55
  element feature maps. Similar operations are repeated in layers
  2,3,4,5. The last two layers are fully connected, taking features
  from the top convolutional layer as input in vector form (6 $\cdot$ 6 $\cdot$
  256 = 9216 dimensions). The final layer is a $C$-way softmax
  function, $C$ being the number of classes. All filters and feature maps are square in shape.}
\label{fig:arch}
\vspace*{-0.3cm}
\end{figure*}

We now describe the large convnet model that will be visualized in
\secc{vis}. The architecture, shown in \fig{arch}, is similar to that
used by \cite{Kriz12} for ImageNet
classification. One difference is that the sparse connections used in
Krizhevsky's layers 3,4,5 (due to the model being split across 2
GPUs) are replaced with dense connections in our model.  Other
important differences relating to layers 1 and 2 were made following
inspection of the visualizations in \fig{compareAlex}, as described in
\secc{selection}.

The model was trained on the ImageNet 2012 training set (1.3 million
images, spread over 1000 different classes).  Each RGB image was
preprocessed by resizing the smallest dimension to 256, cropping the
center 256x256 region, subtracting the per-pixel mean (across all
images) and then using 10 different sub-crops of size 224x224
(corners $+$ center with(out) horizontal flips). Stochastic gradient
descent with a mini-batch size of 128 was used to update the
parameters, starting with a learning rate of $10^{-2}$, in conjunction
with a momentum term of $0.9$. We anneal the
learning rate throughout training manually when the validation error
plateaus. Dropout \cite{Hinton12} is used in the
fully connected layers (6 and 7) with a rate of 0.5. All weights are initialized to $10^{-2}$ and
biases are set to 0.

Visualization of the first layer filters during training reveals that
a few of them dominate, as shown in \fig{compareAlex}(a). To combat
this, we renormalize each filter in the convolutional layers whose RMS
value exceeds a fixed radius of $10^{-1}$ to this fixed radius. This
is crucial, especially in the first layer of the model, where the
input images are roughly in the [-128,128] range.  As in
\cite{Kriz12}, we produce multiple different crops and flips of each
training example to boost training set size. We stopped training after
70 epochs, which took around 12 days on a single
GTX580 GPU, using an implementation based on \cite{Kriz12}.

\section{Convnet Visualization}
\label{sec:vis}
Using the model described in \secc{training}, we now use the deconvnet
to visualize the feature activations on the ImageNet validation set.

\begin{figure*}[h!]
\begin{center}
\includegraphics[width=6.0in]{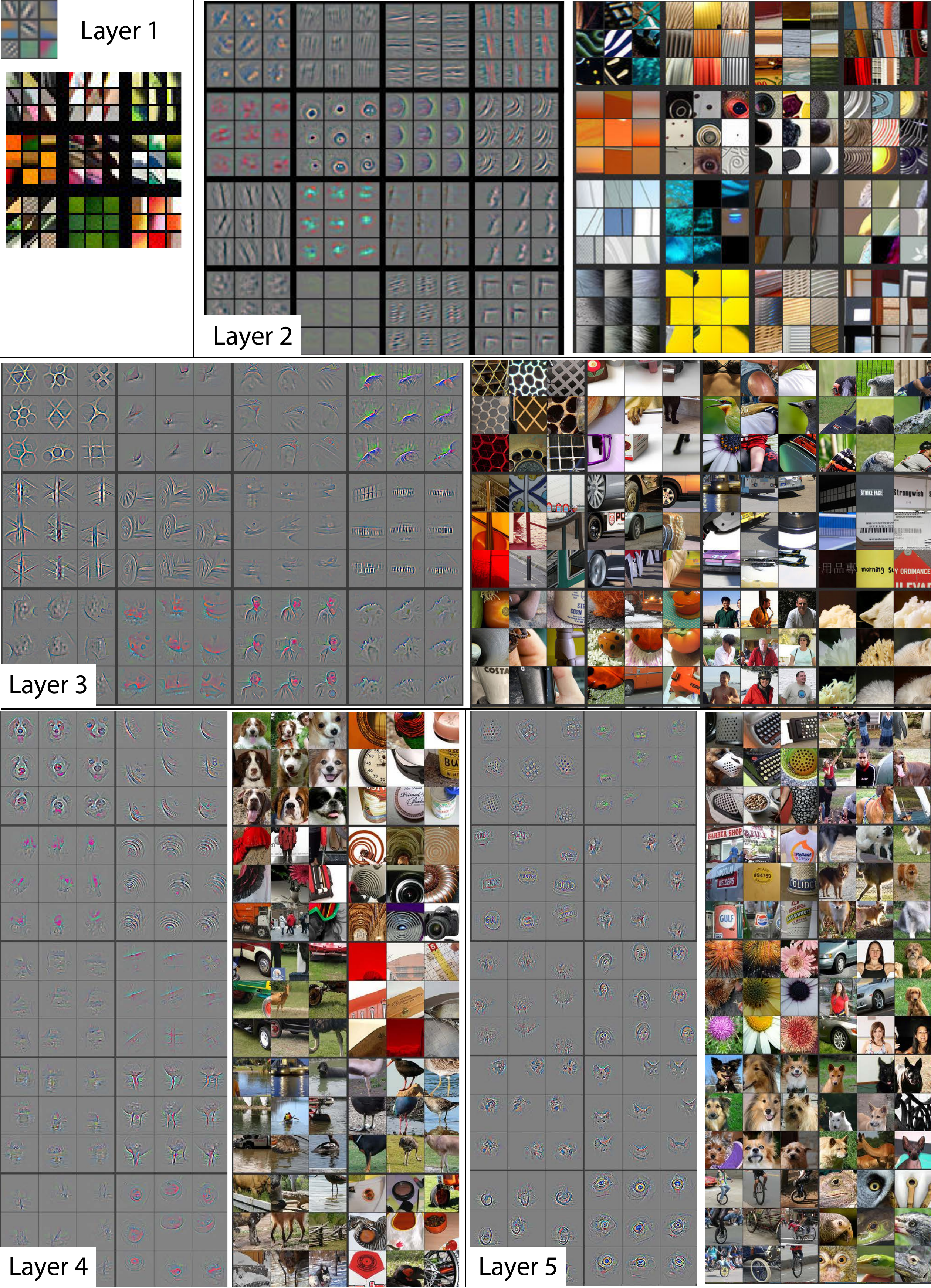}
\end{center}
\vspace*{-0.3cm}
\caption{Visualization of features in a fully trained model. For
  layers 2-5 we show the top 9 activations in a random subset of
  feature maps across the validation data, projected down to pixel
  space using our deconvolutional network approach. Our
  reconstructions are {\em not} samples from the model: they are
  reconstructed patterns from the validation set that cause high activations in a
  given feature map. For each feature map we also show the
  corresponding image patches. Note: (i) the the strong grouping
  within each feature map, (ii) greater invariance at higher layers
  and (iii) exaggeration of discriminative parts of the image,
  e.g.~eyes and noses of dogs (layer 4, row 1, cols 1). Best viewed in electronic form. }
\label{fig:top9feat}
\vspace*{-0.3cm}
\end{figure*}

\noindent {\bf Feature Visualization:} \fig{top9feat} shows feature visualizations from our
model once training is complete. However, instead of showing the
single strongest activation for a given feature map, we show the top 9
activations. Projecting each separately down to pixel space reveals
the different structures that excite a given feature map, hence
showing its invariance to input deformations. Alongside these
visualizations we show the corresponding image patches. These have
greater variation than visualizations as the latter solely focus on
the discriminant structure within each patch. For example, in layer 5,
row 1, col 2, the patches appear to have little in common, but the
visualizations reveal that this particular feature map focuses on the
grass in the background, not the foreground objects.

\onecolumn

\twocolumn

The projections from each layer show the hierarchical nature of
the features in the network. Layer
2 responds to corners and other edge/color conjunctions. Layer 3 has
more complex invariances, capturing similar textures (e.g.~mesh
patterns (Row 1, Col 1); text (R2,C4)). Layer 4 shows significant
variation, but is more class-specific: dog faces
(R1,C1); bird's legs (R4,C2). Layer 5 shows entire objects
with significant pose variation, e.g.~keyboards (R1,C11) and dogs (R4).

\begin{figure*}[t!]
\begin{center}
\includegraphics[width=7in]{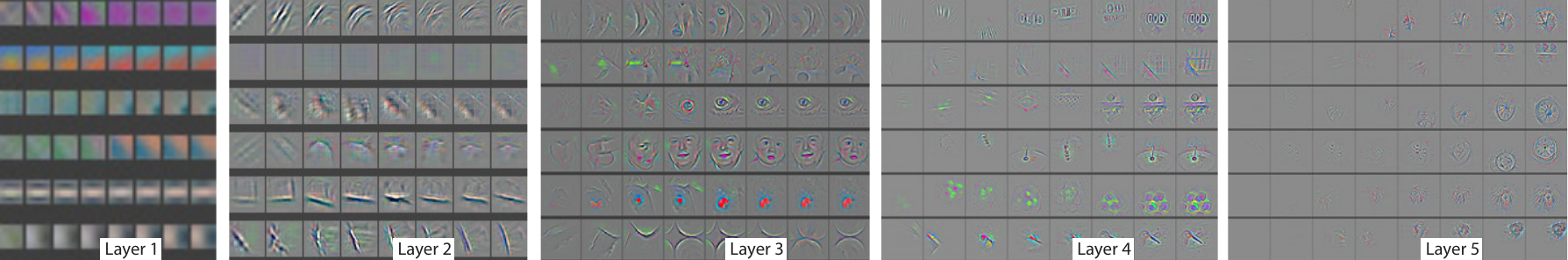}
\end{center}
\vspace*{-0.3cm}
\caption{Evolution of a randomly chosen subset of model features through training. Each layer's
  features are displayed in a different block. Within each
  block, we show a randomly chosen subset of features at epochs [1,2,5,10,20,30,40,64]. The
  visualization shows the strongest activation (across all
  training examples) for a given feature map, projected down to pixel
  space using our deconvnet approach. Color contrast is artificially
  enhanced and the figure is best viewed in
  electronic form.}
\label{fig:evolve}
\vspace*{-0.3cm}
\end{figure*}

\noindent {\bf Feature Evolution during Training:} \fig{evolve}
visualizes the progression during training of the strongest activation
(across all training examples) within a given feature map projected
back to pixel space. Sudden jumps in appearance result from a change
in the image from which the strongest activation originates. The lower
layers of the model can be seen to converge within a few
epochs. However, the upper layers only develop develop after a
considerable number of epochs (40-50), demonstrating the need to let
the models train until fully converged. 

\begin{figure*}[t!]
\begin{center}
\includegraphics[width=6.5in]{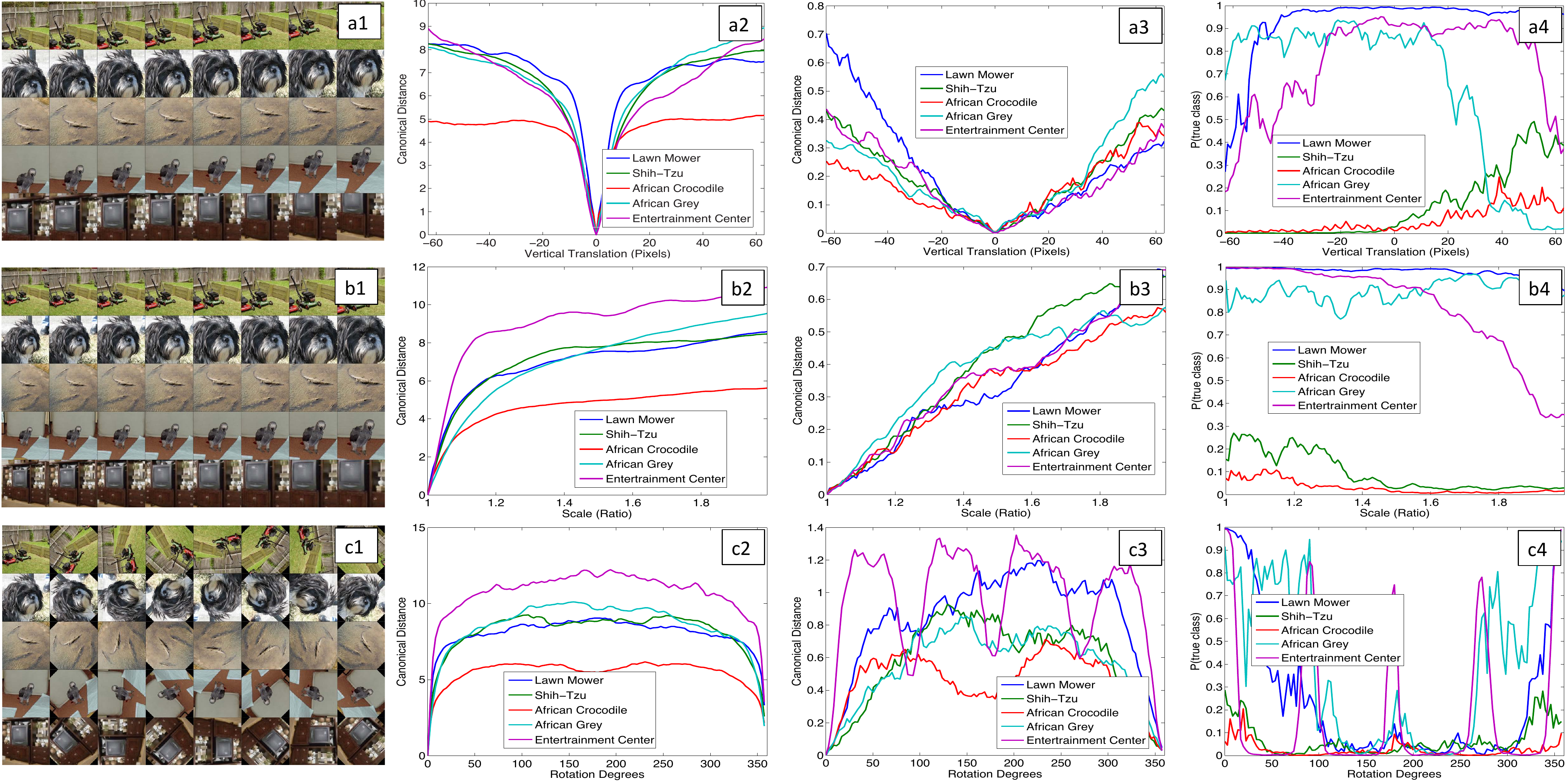}
\end{center}
\vspace*{-0.6cm}
\caption{Analysis of vertical translation, scale, and rotation invariance within the
  model (rows a-c respectively). Col 1: 5 example images undergoing
  the transformations. Col 2 \& 3: Euclidean distance between feature
  vectors from the original and transformed images in layers 1 and 7
  respectively. Col 4: the probability of the true label for each
  image, as the image is transformed. }
\label{fig:invariance}
\vspace*{-0.3cm}
\end{figure*}

\noindent {\bf Feature Invariance:} \fig{invariance} shows 5 sample images being translated, rotated and
scaled by varying degrees while looking at the changes in the feature
vectors from the top and bottom layers of the model, relative to the
untransformed feature. Small
transformations have a dramatic effect in the first layer of the model,
but a lesser impact at the top feature layer, being quasi-linear for
translation \& scaling. The network output is stable to translations
and scalings. In general, the output is not invariant to rotation,
except for object with rotational symmetry (e.g.~entertainment center).



\subsection{Architecture Selection} \label{sec:selection} 
\begin{figure*}[t!]
\begin{center}
\includegraphics[width=6.3in]{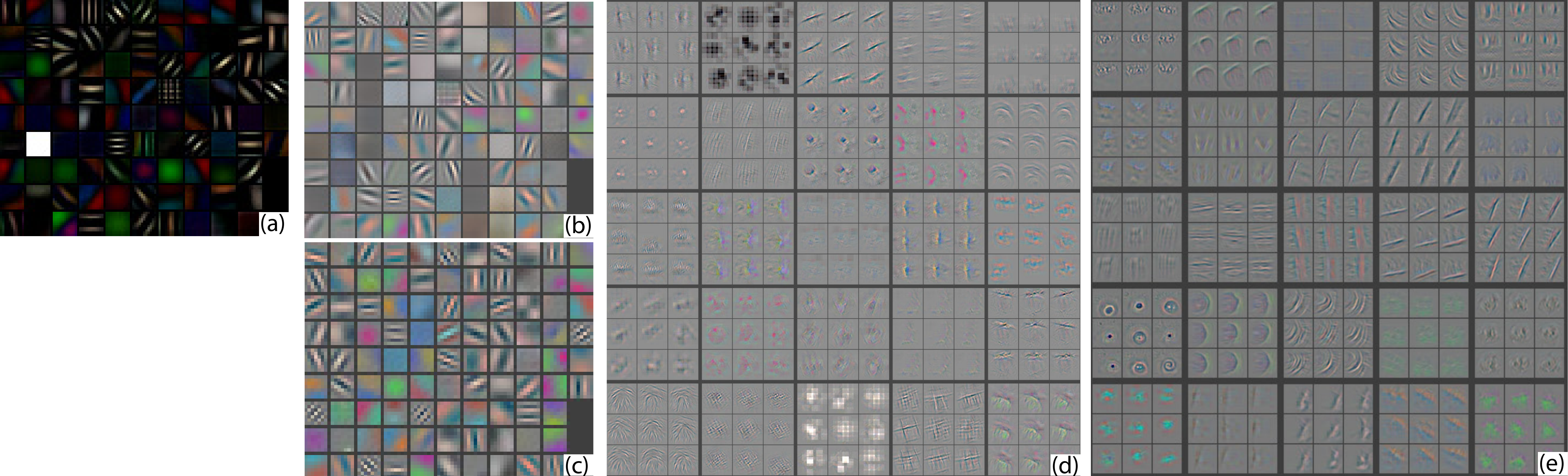}
\end{center}
\vspace*{-0.3cm}
\caption{(a): 1st layer features without feature scale clipping. Note
  that one feature dominates. (b): 1st layer features from
  \cite{Kriz12}. (c): Our 1st layer features. The smaller stride (2 vs
  4) and filter size (7x7 vs 11x11) results in more distinctive
  features and fewer ``dead'' features. (d): Visualizations of 2nd
  layer features from \cite{Kriz12}. (e):
  Visualizations of our 2nd layer features. These are cleaner, with no
  aliasing artifacts that are visible in (d). }
\label{fig:compareAlex}
\vspace*{-0.3cm}
\end{figure*}

While visualization
of a trained model gives insight into its operation, it can also
assist with selecting good architectures in the first place. By
visualizing the first and second layers of Krizhevsky \etal's
architecture (\fig{compareAlex}(b) \& (d)), various problems are
apparent. The first layer filters are a mix of extremely high and low
frequency information, with little coverage of the mid
frequencies. Additionally, the 2nd layer visualization shows aliasing
artifacts caused by the large stride 4 used in the 1st layer
convolutions. To remedy these problems, we (i) reduced the 1st layer
filter size from 11x11 to 7x7 and (ii) made the stride of the
convolution 2, rather than 4. This new architecture retains much more
information in the 1st and 2nd layer features, as shown in
\fig{compareAlex}(c) \& (e). More importantly, it also improves the classification
performance as shown in \secc{modelsizes}.


\subsection{Occlusion Sensitivity}
\begin{figure*}[t!]
\begin{center}
\includegraphics[width=7.0in]{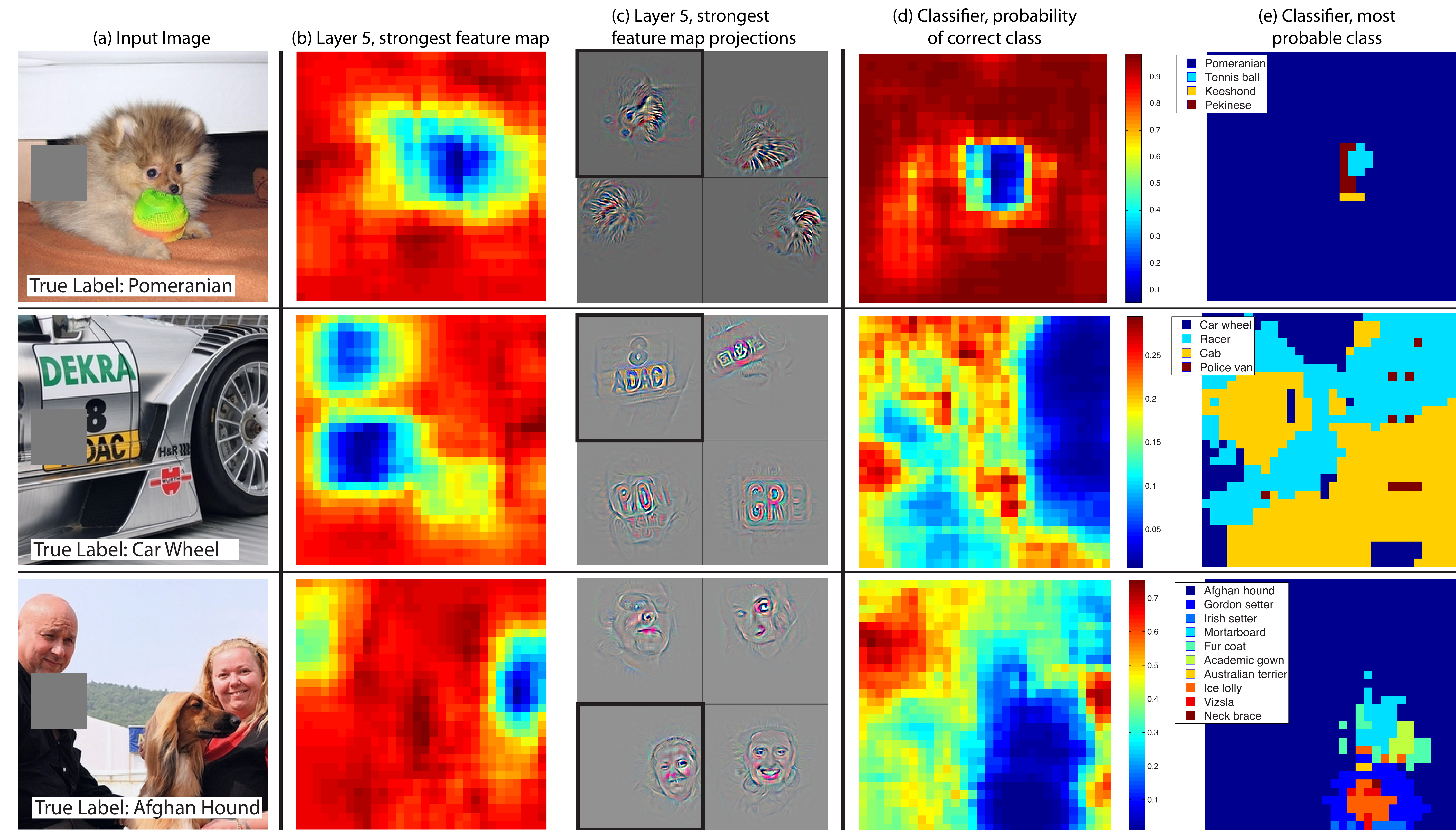}
\end{center}
\vspace*{-0.3cm}
\caption{Three test examples where we systematically cover up different portions
of the scene with a gray square (1st column) and see how the top
(layer 5) feature maps ((b) \& (c)) and classifier output ((d) \& (e)) changes. (b): for each
position of the gray scale, we record the total activation in one
layer 5 feature map (the one with the strongest response in the
unoccluded image). (c): a visualization of this feature map projected
down into the input image (black square), along with visualizations of
this map from other images. The first row example shows the strongest feature
to be the dog's face. When this is covered-up the activity in the
feature map decreases (blue area in (b)). (d): a map of correct class
probability, as a function of the position of the gray
square. E.g.~when the dog's face is obscured, the probability for
``pomeranian'' drops significantly. (e): the most probable label as a
function of occluder position. E.g.~in the 1st row, for most locations
it is ``pomeranian'', but if the dog's face is obscured but not the
ball, then it predicts ``tennis ball''. In the 2nd example, text on
the car is the strongest feature in layer 5, but the classifier is
most sensitive to the wheel. The 3rd example contains multiple
objects. The strongest feature in layer 5 picks out the faces, but the
classifier is sensitive to the dog (blue region in (d)),
since it uses multiple feature maps.}
\label{fig:block_expt}
\vspace*{-0.3cm}
\end{figure*}

With image classification approaches, a natural question is if the
model is truly identifying the location of the object in the image, or
just using the surrounding context. \fig{block_expt} attempts to
answer this question by systematically occluding different portions of
the input image with a grey square, and monitoring the output of the
classifier. The examples clearly show the model is localizing the
objects within the scene, as the probability of the correct class
drops significantly when the object is occluded. \fig{block_expt} also
shows visualizations from the strongest feature map of the top
convolution layer, in addition to
activity in this map (summed over spatial locations) as a function of occluder position. When the
occluder covers the image region that appears in the visualization, we
see a strong drop in activity in the feature map. This shows that the
visualization genuinely corresponds to the image structure that
stimulates that feature map, hence validating the other visualizations
shown in \fig{evolve} and \fig{top9feat}.

\subsection{Correspondence Analysis}
\begin{figure}[t!]
\vspace{-0mm}
\begin{center}
\includegraphics[width=3.2in]{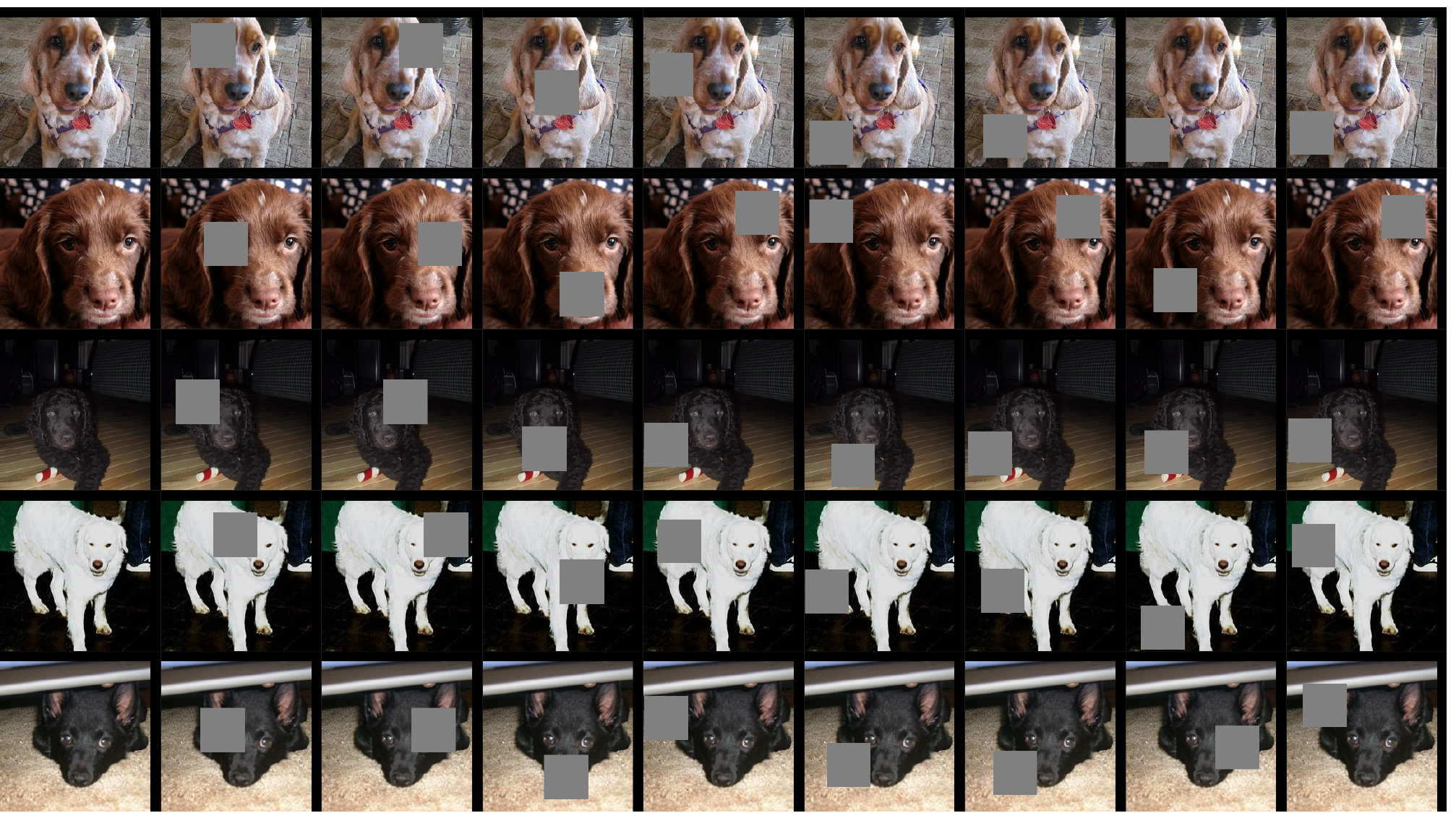}
\end{center}
\vspace*{-0.4cm}
\caption{Images used for correspondence experiments. Col 1: Original
  image. Col 2,3,4: Occlusion of the right eye, left eye, and nose
  respectively. Other columns show examples of random occlusions.}
\label{fig:corr_ims}
\vspace*{-0.0cm}
\end{figure}

\vspace{-2mm}
Deep models differ from many existing recognition approaches in that
there is no explicit mechanism for establishing correspondence between
specific object parts in different images (e.g. faces have a particular
spatial configuration of the eyes and nose). However, an intriguing possibility is that deep models might
be {\em implicitly} computing them. To explore this, we take 5
randomly drawn dog images with frontal pose and systematically mask out
the same part of the face in each image (e.g.~all left eyes, see \fig{corr_ims}). For each
image $i$, we then compute: $\epsilon^l_i = x^l_i - \tilde{x}^l_i$,
where $x^l_i$ and $\tilde{x}^l_i$ are the feature vectors at layer $l$
for the original and occluded images respectively. We then measure the
consistency of this difference vector $\epsilon$ between all related image
pairs $(i,j)$: $\Delta_l  = \sum_{i,j=1, i \neq j}^{5} \mathcal{H}(
\text{sign}(\epsilon^l_i),\text{sign}(\epsilon^l_j))$, where
$\mathcal{H}$ is Hamming distance. A lower value indicates
greater consistency in the change resulting from the masking
operation, hence tighter correspondence between the same object parts
in different images (i.e.~blocking the left eye changes the
feature representation in a consistent way). In \tab{corr} we compare the $\Delta$ score for
three parts of the face (left eye, right eye and nose) to random parts
of the object, using features from layer $l=5$ and $l=7$. The lower score
for these parts, relative to random object regions, for the layer $5$ features
show the model does establish some degree of correspondence.

\begin{table}[!h]
\vspace*{-0mm}
\begin{center}
\small
\begin{tabular}{|l||c|c|}
\hline
& Mean Feature & Mean Feature \\ & Sign Change & Sign Change \\ Occlusion Location & Layer 5 & Layer 7 \\
\hline
Right Eye & $0.067 \pm 0.007 $ & $0.069 \pm 0.015 $\\ \hline
Left Eye & $0.069 \pm 0.007 $ & $0.068 \pm 0.013 $\\ \hline
Nose & $0.079 \pm 0.017 $ & $0.069 \pm 0.011 $\\ \hline \hline
Random & $0.107 \pm 0.017 $ & $0.073 \pm 0.014 $\\ \hline

\end{tabular}
\vspace*{-0mm}
\caption{Measure of correspondence for different object parts in 5
  different dog images. The lower scores for the eyes and nose
  (compared to random object parts) show
  the model implicitly establishing some form of correspondence
  of parts at layer 5 in the model. At layer 7, the scores are more
  similar, perhaps due to upper layers trying to discriminate between the
  different breeds of dog.  }
\label{tab:corr}
\vspace*{-5mm}
\end{center}
\end{table}

\section{Experiments}

\subsection{ImageNet 2012}
\label{ImageNet}
This dataset consists of 1.3M/50k/100k training/validation/test
examples, spread over 1000 categories. \tab{ImageNet} shows our results
on this dataset.

Using the exact architecture specified in \cite{Kriz12}, we attempt to replicate their result on the validation set. We achieve an
error rate within $0.1\%$ of their reported value on the
ImageNet 2012 validation set. 


Next we analyze the performance of our model with the architectural
changes outlined in \secc{selection} ($7\times7$ filters in layer 1 and
stride $2$ convolutions in layers 1 \& 2). This model, shown in
\fig{arch}, significantly outperforms the architecture of
\cite{Kriz12}, beating their single model result by $1.7\%$ (test top-5). When we combine
multiple models, we obtain a test error of {\bf ${\bf 14.8\%}$, the best
published performance on this dataset}\footnote{This performance has
been surpassed in the recent Imagenet 2013 competition (\url{http://www.image-net.org/challenges/LSVRC/2013/results.php}).} (despite only using the 2012 training
set). We note that this error is almost half that of the top non-convnet
entry in the ImageNet 2012 classification challenge, which obtained
$26.2\%$ error \cite{ISI}.

\begin{table}[h!]
\scriptsize
\vspace*{-3mm}
\begin{center}
\begin{tabular}{|l|l|l|l|}
  \hline
          & Val       & Val        & Test \\
 Error \% & Top-1     & Top-5      & Top-5 \\
  \hline \hline
  \cite{ISI}               & -      &  -      & $26.2$ \\ \hline
  \cite{Kriz12}, 1 convnet & $40.7$ &  $18.2$ & $--$ \\
  \cite{Kriz12}, 5 convnets & $38.1$ &  $16.4$ & $16.4$ \\
  \cite{Kriz12}$^*$, 1 convnets & $39.0$ &  $16.6$ & $--$ \\
  \cite{Kriz12}$^*$, 7 convnets & $36.7$ &  $15.4$ & $15.3$ \\
  \hline \hline 
  Our replication of & & & \\
  \cite{Kriz12}, 1 convnet & $40.5$ &  $18.1$ & $--$ \\
  \hline 
  1 convnet as per \fig{arch} & $38.4$ & $16.5$ & $--$ \\
  \hline 
  5 convnets as per \fig{arch} -- (a) & $36.7$ & $15.3$ & $15.3$ \\
  \hline 
  1 convnet as per \fig{arch} but with & & & \\ layers 3,4,5: 512,1024,512
  maps -- (b)& $37.5  $ & $16.0$  & $16.1$ \\
  \hline 
  6 convnets, (a) \& (b) combined & $\bf{36.0}  $ & $\bf{14.7}$  & $\bf{14.8}$ \\
  \hline
\end{tabular}
\vspace*{-2mm}
\caption{ImageNet 2012 classification error rates. The $*$ indicates
  models that were trained on both ImageNet  2011 and 2012 training sets.}
\label{tab:ImageNet}
\vspace*{-3mm}
\end{center}
\end{table}

\noindent {\bf Varying ImageNet Model Sizes:}
 \label{sec:modelsizes} In \tab{modelSizes}, we first explore the
 architecture of \cite{Kriz12} by adjusting the size of layers, or
 removing them entirely. In each case, the model is trained from
 scratch with the revised architecture. Removing the fully connected
 layers (6,7) only gives a slight increase in error. This is surprising, given that they
 contain the majority of model parameters. Removing two of
 the middle convolutional layers also makes a relatively small
 different to the error rate. However, removing both the middle
 convolution layers and the fully connected layers yields a model with
 only 4 layers whose performance is dramatically worse. This would
 suggest that the overall depth of the model is important for
 obtaining good performance. In \tab{modelSizes}, we modify our
 model, shown in \fig{arch}. Changing the size of the fully connected
 layers makes little difference to performance (same for model of
 \cite{Kriz12}). However, increasing the size of the middle convolution layers
 goes give a useful gain in performance. But increasing these, while also
 enlarging the fully connected layers results in over-fitting.

\begin{table}[h!]
\scriptsize
\vspace*{0mm}
\begin{center}
\begin{tabular}{|l|l|l|l|}
  \hline
  & Train    & Val        & Val         \\
Error \%  & Top-1    & Top-1      & Top-5       \\ \hline
\hline 
Our replication of & & & \\
\cite{Kriz12}, 1 convnet & $35.1$ & $40.5$ & $18.1$  \\
\hline 
Removed layers 3,4 & $41.8  $ & $45.4  $ & $22.1  $ \\
\hline 
Removed layer 7 & $27.4  $ & $40.0$ & $18.4  $ \\
\hline 
Removed layers 6,7 & $27.4  $ & $44.8  $ & $22.4  $ \\
\hline
Removed layer 3,4,6,7 & $71.1$ & $71.3$ & $50.1$ \\
\hline 
Adjust layers 6,7: 2048 units & $40.3  $ & $41.7  $ & $18.8  $ \\
\hline 
Adjust layers 6,7: 8192 units & $26.8  $ & $40.0$  & $18.1$ \\
\hline \hline \hline 
Our Model (as per \fig{arch}) & $33.1  $ & $38.4  $ & $16.5  $ \\
\hline 
Adjust layers 6,7: 2048 units & $38.2  $ & $40.2  $ & $17.6  $ \\
\hline 
Adjust layers 6,7: 8192 units & $22.0  $ & $38.8  $ & $17.0  $ \\
\hline 
Adjust layers 3,4,5: 512,1024,512 maps & $18.8  $ & $\bf{37.5}  $ & $\bf{16.0}  $ \\
\hline 
Adjust layers 6,7: 8192 units and & & & \\ Layers 3,4,5: 512,1024,512 maps & $\bf{10.0}  $ & $38.3  $ & $16.9  $ \\
  \hline
\end{tabular}
\vspace*{-2mm}
\caption{ImageNet 2012 classification error rates with various
  architectural changes to the model of \cite{Kriz12} and our model
  (see \fig{arch}). }
\label{tab:modelSizes}
\vspace*{-8mm}
\end{center}
\end{table}

\subsection{Feature Generalization}
\vspace*{-2mm}
The experiments above show the importance of the convolutional part of
our ImageNet model in obtaining state-of-the-art performance.
This is supported by the visualizations of \fig{top9feat} which show
the complex invariances learned in the convolutional layers. We now
explore the ability of these feature extraction layers to generalize
to other datasets, namely Caltech-101 \cite{caltech101}, Caltech-256
\cite{caltech256} and PASCAL
VOC 2012. To do this, we keep layers 1-7 of our ImageNet-trained model
fixed and train a new softmax classifier on top (for the appropriate
number of classes) using the training images of the new dataset. Since
the softmax contains relatively few parameters, it can be trained
quickly from a relatively small number of examples, as is the case for
certain datasets.

The classifiers used by our model (a softmax) and other approaches
(typically a linear SVM) are of similar complexity, thus the
experiments compare our feature representation, learned from ImageNet,
with the hand-crafted features used by other methods. It is important
to note that {\em both} our feature representation and the
hand-crafted features are designed using images beyond the Caltech and
PASCAL training sets. For example, the hyper-parameters in HOG
descriptors were determined through systematic experiments on a
pedestrian dataset \cite{Dalal05}. 
We also try a second strategy of training a model from scratch,
i.e.~resetting layers 1-7 to random values and train them, as well as
the softmax, on the training images of the dataset.

One complication is that some of the Caltech datasets have some images that are
also in the ImageNet training data. Using normalized correlation, we
identified these few ``overlap'' images\footnote{ For
Caltech-101, we found 44 images in common (out of 9,144 total images),
with a maximum overlap of 10 for any given class. For Caltech-256, we
found 243 images in common (out of 30,607 total images), with a
maximum overlap of 18 for any given class.} and removed them from our
Imagenet training set and then retrained our Imagenet models, so avoiding the
possibility of train/test contamination.  

\vspace{3mm}
\noindent {\bf Caltech-101:} We follow the procedure of
\cite{caltech101} and randomly select 15 or 30 images per class for
training and test on up to 50 images per class reporting the average
of the per-class accuracies in \tab{caltech101}, using 5 train/test
folds. Training took 17 minutes for 30 images/class.  The
pre-trained model beats the best reported result for 30 images/class
from \cite{Bo13} by 2.2\%. The convnet model trained from
scratch however does terribly, only achieving 46.5\%.




\begin{table}[h!]
\small
\vspace*{-3mm}
\begin{center}
\begin{tabular}{|l|l|l|}
  \hline
 & Acc \%      & Acc \% \\
 \# Train& 15/class  & 30/class \\
  \hline
  \cite{Bo13} & $-$ & $81.4 \pm 0.33$ \\
  \cite{Yang09} & $73.2$ & $84.3 $ \\
  \hline \hline %
\hline 
Non-pretrained convnet & $22.8 \pm 1.5$ & $46.5 \pm 1.7$ \\
\hline 
ImageNet-pretrained convnet & $\bf{83.8 \pm 0.5}$ & $\bf{86.5 \pm 0.5}$ \\

  \hline
\end{tabular}
\vspace*{-3mm}
\caption{Caltech-101 classification accuracy for our convnet models,
  against two leading alternate approaches.}
\label{tab:caltech101}
\vspace*{-3mm}
\end{center}
\end{table}

\begin{figure}[t!]
\begin{center}
\includegraphics[width=2.5in]{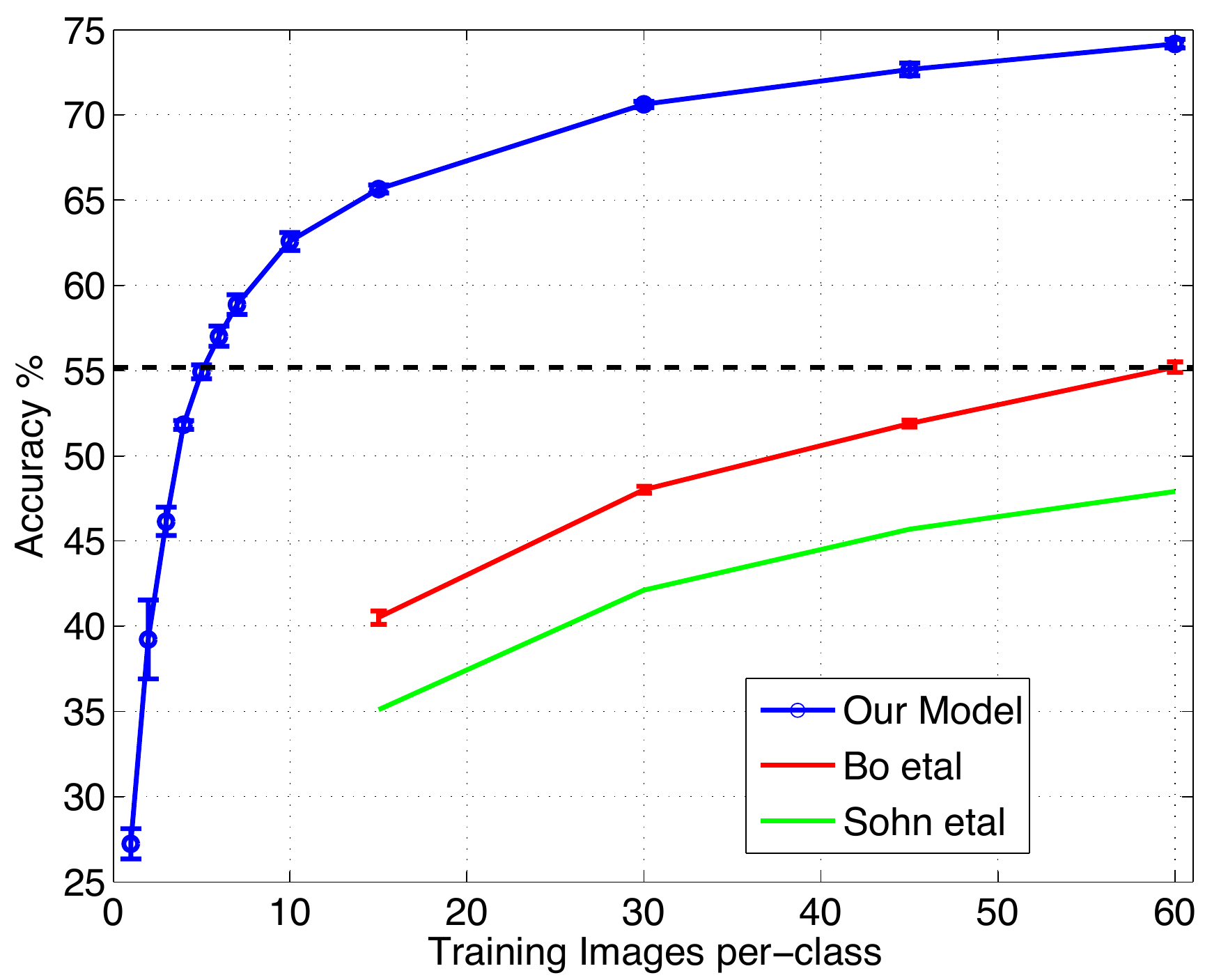}
\end{center}
\vspace*{-0.3cm}
\caption{Caltech-256 classification performance as the number of
  training images per class is varied. Using only 6 training examples
  per class with our pre-trained feature extractor, we surpass best
  reported result by \cite{Bo13}. }
\label{fig:256plot}
\vspace*{-0.4cm}
\end{figure}

\noindent {\bf Caltech-256:} We follow the procedure of
\cite{caltech256}, selecting 15, 30, 45, or 60 training images per
class, reporting the average of the per-class accuracies in
\tab{caltech256}. Our ImageNet-pretrained model beats the current
state-of-the-art results obtained by Bo \etal \cite{Bo13} by a
significant margin: 74.2\% vs 55.2\% for 60 training
images/class. However, as with Caltech-101, the model trained from
scratch does poorly. In \fig{256plot}, we explore the ``one-shot
learning'' \cite{caltech101} regime. With our pre-trained model, just
6 Caltech-256 training images are needed to beat the leading method
using 10 times as many images. This shows the power of the ImageNet
feature extractor.
\begin{table}[h!]
\small
\vspace*{0mm}
\begin{center}
\tabcolsep=0.07cm
\begin{tabular}{|l|l|l|l|l|}
  \hline
  & Acc \%    & Acc \%      & Acc \%     & Acc \%\\
\# Train  & 15/class & 30/class   & 45/class & 60/class \\
  \hline
  \cite{Sohn11} & $35.1$ &  $42.1 $ & $45.7$ & $47.9$ \\
  \cite{Bo13} & $40.5 \pm 0.4$ &  $48.0 \pm 0.2$ & $51.9 \pm 0.2$ & $55.2 \pm 0.3$ \\
  \hline \hline %

\hline 
Non-pretr. & $9.0 \pm 1.4$ & $22.5 \pm 0.7$ & $31.2 \pm 0.5$ & $38.8 \pm 1.4$ \\
\hline 
ImageNet-pretr. & $\bf{65.7 \pm 0.2}$ & $\bf{70.6 \pm 0.2}$ & $\bf{72.7 \pm 0.4}$ & $\bf{74.2 \pm 0.3}$ \\
  \hline
\end{tabular}
\vspace*{-3mm}
\caption{Caltech 256 classification accuracies.}
\label{tab:caltech256}
\vspace*{-4mm}
\end{center}
\end{table}

\vspace{2mm}
\noindent {\bf PASCAL 2012:} We used the standard training and validation
images to train a 20-way softmax on top of the ImageNet-pretrained
convnet. This is not ideal, as PASCAL images can contain multiple
objects and our model just provides a single exclusive prediction for each
image. \tab{pascal} shows the results on the test set. The PASCAL
and ImageNet images are quite different in nature, the former being
full scenes unlike the latter. This may explain our
mean performance being $3.2\%$ lower than the leading \cite{nus}
result, however we do beat them on 5 classes, sometimes by large margins.


\begin{table}[!h]
\tiny
\begin{center}
\begin{tabular}{|l|l|l|l||l|l|l|l|}
\hline
Acc \% & [A] & [B] & Ours & Acc \% & [A] & [B] & Ours \\ \hline
\hline
Airplane& 92.0 & {\bf 97.3} & 96.0 &Dining tab &63.2 &  {\bf 77.8}&67.7 \\ \hline
Bicycle & 74.2 & {\bf 84.2} & 77.1 &Dog& 68.9 & 83.0 &  {\bf 87.8} \\ \hline
Bird& 73.0 & 80.8 & {\bf 88.4}& Horse & 78.2 &  {\bf 87.5} & 86.0  \\ \hline
Boat & 77.5 & 85.3 & {\bf 85.5}&Motorbike & 81.0 &  {\bf 90.1} & 85.1 \\ \hline
Bottle& 54.3 & {\bf 60.8} & 55.8&Person & 91.6 &  {\bf 95.0} & 90.9  \\ \hline
Bus& 85.2 & {\bf 89.9} & 85.8& Potted pl & 55.9 &  {\bf 57.8} & 52.2\\ \hline
Car&81.9&{\bf 86.8} & 78.6&Sheep & 69.4 & 79.2 &  {\bf 83.6} \\ \hline
Cat&76.4 &89.3 & {\bf 91.2}& Sofa & 65.4 &  {\bf 73.4} & 61.1 \\ \hline
Chair&65.2&{\bf 75.4}& 65.0&Train & 86.7 &  {\bf 94.5} & 91.8  \\ \hline
Cow&63.2&{\bf 77.8}& 74.4& Tv & 77.4 &  {\bf 80.7} & 76.1  \\ \hline \hline
Mean & 74.3 & {\bf 82.2} & 79.0 & \# won & 0 & {\bf 15} &5 \\ \hline
\end{tabular}
\vspace*{-3mm}
\caption{PASCAL 2012 classification results, comparing our
  Imagenet-pretrained convnet against the leading two methods ([A]=
  \cite{cvc} and [B] = \cite{nus}). }
\label{tab:pascal}
\vspace*{-4mm}
\end{center}
\end{table}


\subsection{Feature Analysis} \label{sec:pretrain} We explore how
discriminative the features in each layer of our Imagenet-pretrained
model are. We do this by varying the number of layers retained from
the ImageNet model and place either a linear SVM or softmax classifier
on top. \tab{supretrain} shows results on Caltech-101 and
Caltech-256. For both datasets, a steady improvement can be seen as we
ascend the model, with best results being obtained by using all
layers. This supports the premise that as the feature hierarchies
become deeper, they learn increasingly powerful features.


\begin{table}[h!]
\small
\vspace*{-2mm}
\begin{center}
\tabcolsep=0.11cm
\begin{tabular}{|l|l|l|}
  \hline

  & Cal-101 & Cal-256  \\
  & (30/class) & (60/class)  \\
\hline 
SVM (1) & $44.8 \pm 0.7$ & $24.6 \pm 0.4$ \\
\hline 
SVM (2) & $66.2 \pm 0.5$ & $39.6 \pm 0.3$ \\
\hline 
SVM (3) & $72.3 \pm 0.4$ & $46.0 \pm 0.3$ \\
\hline 
SVM (4) & $76.6 \pm 0.4$ & $51.3 \pm 0.1$ \\
\hline 
SVM (5) & $\bf{86.2 \pm 0.8}$ & $65.6 \pm 0.3$ \\
\hline 
SVM (7) & $\bf{85.5 \pm 0.4}$ & $\bf{71.7 \pm 0.2}$ \\
\hline 
Softmax (5) & $82.9 \pm 0.4$ & $65.7 \pm 0.5$ \\
\hline 
Softmax (7) & $\bf{85.4 \pm 0.4}$ & $\bf{72.6 \pm 0.1}$ \\


  \hline
\end{tabular}
\vspace*{0mm}
\caption{Analysis of the discriminative information contained in each
  layer of feature maps within our ImageNet-pretrained convnet. We
  train either a linear SVM or softmax on features from different
  layers (as indicated in brackets) from the convnet. Higher layers generally
  produce more discriminative features.}
\label{tab:supretrain}
\vspace*{-8mm}
\end{center}
\end{table}



\section{Discussion}
We explored large convolutional neural network models, trained for
image classification, in a number ways. First, we presented a novel
way to visualize the activity within the model. This reveals the
features to be far from random, uninterpretable patterns. Rather, they
show many intuitively desirable properties such as compositionality,
increasing invariance and class discrimination as we ascend the
layers. We also showed how these visualization can be used to debug
problems with the model to obtain better results, for example
improving on Krizhevsky \etal's \cite{Kriz12} impressive ImageNet 2012
result. We then demonstrated through a series of occlusion experiments that
the model, while trained for classification, is highly sensitive to
local structure in the image and is not just using broad scene
context. An ablation study on the model revealed that having a minimum
depth to the network, rather than any individual section, is vital to
the model's performance.

Finally, we showed how the ImageNet trained model can generalize
well to other datasets. For Caltech-101 and Caltech-256,
the datasets are similar enough that we can beat the best reported
results, in the latter case by a significant margin.
This result brings into question to utility of benchmarks with small
(i.e.~$<10^4$) training sets. Our convnet model generalized less well
to the PASCAL data, perhaps suffering from dataset bias
\cite{Torralba11}, although it was still within $3.2\%$ of the best reported
result, despite no tuning for the task. For example, our performance
might improve if a different loss function was used that permitted
multiple objects per image. This would naturally enable the networks
to tackle the object detection as well.




\section*{Acknowledgments}
The authors are very grateful for support by NSF grant IIS-1116923, Microsoft Research
and a Sloan Fellowship.





\bibliography{demystify}
\bibliographystyle{icml2013}

\end{document}